\documentclass[preprint]{elsarticle}
\makeatletter
\def\ps@pprintTitle{%
 \let\@oddhead\@empty
 \let\@evenhead\@empty
 \def\@oddfoot{\centerline{\thepage}}%
 \let\@evenfoot\@oddfoot}
\makeatother

\usepackage{amssymb}
\usepackage{amsthm}

\usepackage{lineno}

\usepackage{xspace}
\usepackage{xcolor}
\usepackage{amsmath}
\usepackage{amsfonts}
\usepackage{url}

\usepackage{multirow}
\usepackage{subcaption}

\usepackage{graphicx}
\newcommand{\etal}{\textit{et al}.\@\xspace}
\newcommand{\eg}{\textit{e}.\textit{g}.\,\@\xspace}

\newcommand{\rev}[1]{{#1}} % UNCOMMENT THIS LINE TOO

\journal{Pattern Recognition}

\begin{document}

\begin{frontmatter}

%% Title, authors and addresses

%% use the tnoteref command within \title for footnotes;
%% use the tnotetext command for theassociated footnote;
%% use the fnref command within \author or \address for footnotes;
%% use the fntext command for theassociated footnote;
%% use the corref command within \author for corresponding author footnotes;
%% use the cortext command for theassociated footnote;
%% use the ead command for the email address,
%% and the form \ead[url] for the home page:
%% \title{Title\tnoteref{label1}}
%% \tnotetext[label1]{}
%% \author{Name\corref{cor1}\fnref{label2}}
%% \ead{email address}
%% \ead[url]{home page}
%% \fntext[label2]{}
%% \cortext[cor1]{}
%% \address{Address\fnref{label3}}
%% \fntext[label3]{}

\title{SSP-Net: Scalable Sequential Pyramid Networks\\ for Real-Time 3D Human Pose Regression}

%% use optional labels to link authors explicitly to addresses:
%% \author[label1,label2]{}
%% \address[label1]{}
%% \address[label2]{}

\author[etis]{Diogo Carbonera Luvizon\corref{mycorrespondingauthor}}

\cortext[mycorrespondingauthor]{Corresponding author}
\ead{diogo.luvizon@ensea.fr}

\author[etis,ibisc]{Hedi Tabia}

\author[etis,ligm]{David Picard}

\address[etis]{ETIS UMR 8051, Paris Seine University, ENSEA, CNRS, F-95000, Cergy, France}
\address[ibisc]{IBISC, Univ. d\'{}Evry Val d\'{}Essonne, Universit\'{e} Paris Saclay}
\address[ligm]{LIGM, UMR 8049, \'{E}cole des Ponts, UPE, Champs-sur-Marne, France}

\begin{abstract}
  % REWRITE THE ABSTRACT
  \rev{
  In this paper we propose a highly scalable \rev{convolutional neural network},
  end-to-end trainable, for
  real-time 3D human pose regression from still RGB images.
  We call this approach the Scalable Sequential Pyramid Networks
  (SSP-Net) as it is trained with refined supervision at multiple scales in a sequential manner.
  Our network requires a single training procedure and is capable of
  producing its best predictions at 120 frames per second (FPS), or acceptable predictions at
  more than 200 FPS when cut at test time.
  We show that the proposed regression approach is invariant to the size of feature maps,
  allowing our method to perform multi-resolution intermediate supervisions and
  reaching results comparable to the state-of-the-art with very low resolution
  feature maps.}
  We demonstrate the accuracy and the effectiveness of our method by providing
  extensive experiments on two of the most important publicly available
  datasets for 3D pose estimation, Human3.6M and MPI-INF-3DHP.
  Additionally, we provide relevant insights about our decisions on the network
  architecture and show its flexibility to meet the best precision-speed
  compromise.
\end{abstract}


\end{frontmatter}

%% \linenumbers

%% main text

%% The Appendices part is started with the command \appendix;
%% appendix sections are then done as normal sections
%% \appendix

%% \section{}
%% \label{}

%%%%%%%%%%%%%%%%%%%%%%%%%%%%%%%%%%%%%%%%%%%%%%%%%%%%%%%%%%%%%%%%%%%%%%%%%%%%%%
%%%%%%% INTRO
%%%%%%%%%%%%%%%%%%%%%%%%%%%%%%%%%%%%%%%%%%%%%%%%%%%%%%%%%%%%%%%%%%%%%%%%%%%%%%

\section{Introduction}
\label{sec:introduction}

\rev{
Predicting 3D human poses from monocular images is an important task
that benefits several applications, from human understanding and action
recognition~\cite{Luvizon_2018_CVPR} to human shape analysis and character
control~\cite{PISHCHULIN2017276}, among many others.
As a consequence of its high relevance, 3D human pose estimation is a very active topic,
also due to the several challenges involved,
such as the complexity in the human body structure, the variations
in the visual aspects from one person to another, and the possibility of
one or more body parts being occluded in the images.
To handle these challenging cases, multi-scale analysis is traditionally
used to allow a multi-level scene understanding.
}
%
%In general, human pose estimation can be seen from two different perspectives,
%namely as a correlated part detection problem or as a regression problem.
%
%Detection based approaches commonly try to detect body joints individually,
%which are aggregated a posteriori to form one pose prediction.  In contrast,
%methods based on regression use a function to map directly input images to
%body joint positions.

% Change the motivation here from "softargmax" to "multi-level supervision", which
% is possible thanks to the softargmax.
With the breakthrough of deep Convolutional Neural Networks
(CNNs)~\cite{Toshev_CVPR_2014} alongside consistent computational power
increase, human pose estimation methods have shifted from classical
approaches~\cite{Pishchulin_ICCV_2013, Ladicky_CVPR_2013} to deep
architectures~\cite{VNect_SIGGRAPH2017, Chen_2017_CVPR}.
%
%In many recent works from different domains, CNN based methods have overcome
%classical approaches by a large margin~\cite{He_2016_CVPR, Simonyan15}.  
%A key benefit from CNN is that the full pipeline is differentiable, allowing
%end-to-end learning.
%
%However, recent detection methods for 2D and 3D pose
%estimation~\cite{yang2017pyramid, ChenSWLY17, Pavlakos_2017_CVPR} are not fully
%differentiable due to the required argmax.
%
%Additionally, the precision of predicted body joints is proportional to that of
%the heatmap resolution, which leads such approaches to high memory consumption
%and high computational requirements.
\rev{Most of current deep learning approaches for 3D human pose estimation are based on an extension of the stacked hourglass model \cite{Newell_ECCV_2016} where each body joint is associated with a volumetric heatmap (\emph{e.g.}, \cite{Sun_2018_ECCV}) that corresponds to the probability density of the joint in the 3D space. These volumetric heatmaps have two main issues. First, the accuracy of the pose estimation is very sensitive to the resolution of the volumetric heatmap, since the precision of the prediction is directly related to the volume of space encoded by a single voxel of the heatmap. Second, as large volumetric heatmaps are preferred, these methods require large amounts of memory to store the activations.
The combination of these two issues results in neural architectures that do not scale well, that is, they are either accurate but very slow or quick but inaccurate. Furthermore, a new model has to be trained specifically for the chosen trade-off between speed and accuracy.}

%An alternative is to perform coordinates regression directly from
%images~\cite{Toshev_CVPR_2014}.
%However, due to the high variance in both images and in the high articulated
%body skeleton, simple regression approaches such as fully connected layers are
%usually sub-optimized, generally resulting in lower precision if compared to
%detection approaches.
%A solution to this problem is to use the differentiable expectancy of
%normalized heatmaps instead of the non differentiable argmax.
%Another advantage of regression approaches is the easy combination of 3D and
%2D annotated data just by not propagating the error on $z$ in the last case,
%which has been proved a very efficient data augmentation technique.

% Four claims is too much. We should claim:
% 1) A new network architecture.
% 2) Dense supervision on multi-level predictions
%Despite the recent progress, current methods still depend on expensive
%3D heatmaps~\cite{Sun_2018_ECCV}. Moreover, the popular hourglass
%networks~\cite{Newell_ECCV_2016}, very common on pose estimation methods,
%are not designed to consider predictions at multiple scales.
\rev{In the light of the limitations of current methods, we propose a new neural architecture that solves the scalability issues, by regressing the pose in multiple scales in a sequential coarse-to-fine approach. 
We call this approach the \emph{Scalable Sequential Pyramid Networks} (SSP-Net).
With a single training procedure, the SSP-Net produces a full model with several refined prediction outputs that can be cut a test time to select the best accuracy \emph{vs} speed trade off.
The contribution of this paper is an extremely fast 3D human body pose estimation architecture that obtains state of the art results at over 100 FPS.
We also show that our method is robust to the resolution of the model, as it is able to obtain subpixel accuracy, leading to competitive results even for $4\times 4$ pixels feature maps.}
%resulting in the following
% contributions:
% \textit{first}, we propose the Scalable Sequential Pyramid Networks, which is
% a new, fast, and efficient CNN architecture, which produces highly precise 3D
% human pose predictions at speed ranging from 120 FPS up to 290 FPS with a single training procedure.
% \textit{Second}, we are the first to propose dense re-injected supervision and
% dense connections at all scales, leading to very successful results both in
% accuracy and speed.
% \textit{Third}, we present a new 3D pose regression approach that gives
% state-of-the-art results on 3D human pose estimation without
% requiring expensive 3D heatmaps,
% and \textit{fourth}, the proposed regression approach does not require any
% parametrization and is invariant to the feature maps resolution, resulting in
% sub-pixel accuracy and competitive results even with feature maps of $4\times4$
% pixels.

The rest of this paper is divided as follows. In the next section, we present a
review of the related work. The architecture of the proposed
network and the proposed regression method are presented in
Section~\ref{sec:method}. In Section~\ref{sec:experiments}, we show the
experimental evaluation of our method on 3D human pose estimation, providing
intuitions on the proposed method based on a detailed ablation
study. We conclude the paper in Section~\ref{sec:conclusion}.

%%%%%%%%%%%%%%%%%%%%%%%%%%%%%%%%%%%%%%%%%%%%%%%%%%%%%%%%%%%%%%%%%%%%%%%%%%%%%%
%%%%%%% RELATED WORK
%%%%%%%%%%%%%%%%%%%%%%%%%%%%%%%%%%%%%%%%%%%%%%%%%%%%%%%%%%%%%%%%%%%%%%%%%%%%%%

\section{Related work}
\label{sec:related-work}

In this section, we review some of the recent methods most related to our work,
which are divide into two groups: \textit{3D human pose estimation} and
\textit{Multi-stage architectures for human pose estimation}.
We encourage readers to read the survey on 3D human pose
estimation in~\cite{SARAFIANOS20161} for a more detailed bibliographic review.

\subsection{3D human pose estimation}

% Why detection approaches are not common for 3D pose estimation and
% why regression approaches are better than detection?
Estimating the human body joints in 3D coordinates from monocular RGB
images is a challenging problem with a vast bibliography available in the
literature~\cite{ATREVI2017389, Ionesc_ICCV_2011, Moll_CVPR_2014,
TekinKSLF16, Li_2015_ICCV, Ionescu_CVPR_2014}.
Despite the fact that methods for 2D pose estimation are mainly based on 
detection~\cite{Newell_ECCV_2016, insafutdinov17cvpr}, 3D pose estimation is
frequently handled as a regression problem~\cite{Agarwal, zhou2016deep}.
\rev{
The main reason is due to the additional third dimension in 3D predictions,
which significantly increases the required memory and computations,
especially in detection based approaches, were the space is frequently
represented by voxels~\cite{Pavlakos_2017_CVPR}.
On the other hand, regression methods handle the problem
more efficiently, usually resulting in precise estimations with lower resolution~\cite{Luvizon_2018_CVPR}.
}

A common approach for 3D human pose \rev{regression} is to lift 3D coordinates from
2D predictions~\cite{Popa_CVPR_2017, Martinez_2017}.
\rev{
Despite being robust to visual variations, lifting 3D poses from 2D points
is an ill-defined problem, which can result in ambiguity.
In that case, incoherent predictions are common, which requires a matching strategy between the estimated 3D
poses and a structural model~\cite{Chen_2017_CVPR}.
As an alternative, the body joints can be represented relative
to their parent joints, requiring the prediction of the delta between two
neighbour joints~\cite{Sun_2017_ICCV, luo2018orinet}.
}
This approach reduces the variance in the target space. However,
it introduces an accumulative error propagated from the root joint to
the body extremities.

Another problem related to 3D pose estimation is the lack of rich visual data.
Since precise 3D annotations depend on expensive and complex Motion
Capture (MoCap) systems, public datasets are usually collected in controlled
environment with static and clean background, despite having few subjects.
To alleviate this problem, Mehta \etal~\cite{Mehta_2017_3DV} proposed to
first train a 2D model on data collected ``in-the-wild'' with 2D manual
annotations, and then to use transfer learning to build a neural network that
predicts 3D joint positions with respect to the root joint.
\rev{Transfer learning is an useful but tricky technique.
For that reason, in our approach, we decided to have a single training
procedure that uses manually 2D annotated data in-the-wild simultaneously
with high precise MoCap data.}

\subsection{Multi-stage architectures for human pose estimation}

Multi-stage architectures have been widely used for human pose estimation,
specially for the more established problem of 2D pose
estimation~\cite{Wei_CVPR_2016},
\rev{usually as sequential predictions~\cite{Gkioxari_ECCV_2016} or by means
of recurrent networks~\cite{Hu_2016_CVPR}}.
A common practice in previous methods is to regress heatmap representations
\rev{corresponding to a map of scores for a given body joint.}
The refinement of such heatmaps is crucial for achieving good precision, as
noted in \cite{Bulat_ECCV_2016}.
Following this idea, Newell \etal~\cite{Newell_ECCV_2016} proposed the
stacked hourglass architecture, which is essentially a sequence of U-nets,
each one producing a new set of heatmaps that are
refined by further hourglasses.
\rev{Approaches based on heatmap estimation have two drawbacks}:
\textit{first}, predicted heatmaps require \rev{an elevated}
resolution for acceptable precision, since the body
joint coordinates are extracted in a post-processing stage based on the
argument of the maximum a posteriori probability (MAP or argmax).
The \textit{second} limitation is the requirement for
artificially generated ground truth heatmaps during training,
since argmax is not differentiable.
\rev{Contrarily, in our method we can have precise predictions with very low resolution feature maps, in addition to not requiring artificially generated ground truth.}

On 3D scenarios, Zhou \etal~\cite{Zhou_2017_ICCV} benefits from 2D heatmaps
to guide 3D pose regression, introducing a weakly-supervised approach for
lifting 3D predictions from 2D data, and Pavlakos
\etal~\cite{Pavlakos_2017_CVPR} extended the Staked Hourglass network to
volumetric heatmaps prediction, on which the $z$ coordinate is encoded in the
additional heatmap dimension.
However, \rev{the method proposed in \cite{Pavlakos_2017_CVPR}} suffers
from the significant increase in the number of
parameters and in the required memory to store all the intermediate values,
due to the highly expensive volumetric heatmaps.
This problem can be alleviated by the differentiable version of
argmax~\cite{Luvizon_2017_CoRR, Luvizon_2018_CVPR}, also
called \textit{integral regression} in \cite{Sun_2018_ECCV}, but it remains
dependent on a costly voxilized representation of the 3D space.

The method presented in this work differs from all previous approaches in
several aspects. \textit{First}, it departs from requiring volumetric
representations by predicting pairs of heatmaps and depth maps.
\textit{Second}, differently from the stacked hourglass architecture, our
method has intermediate supervision at different scales, providing
different levels of semantic and resolution, which are all aggregated in a
densely connected way for better predictions refinement. \textit{Third},
after a single training procedure, our scalable network can be cut at different
positions, providing a vast trade off for precision \textit{vs.} speed.
All these advantages result from the proposed architecture, as detailed next.

%%%%%%%%%%%%%%%%%%%%%%%%%%%%%%%%%%%%%%%%%%%%%%%%%%%%%%%%%%%%%%%%%%%%%%%%%%%%%%
%%%%%%% METHOD
%%%%%%%%%%%%%%%%%%%%%%%%%%%%%%%%%%%%%%%%%%%%%%%%%%%%%%%%%%%%%%%%%%%%%%%%%%%%%%

\section{Scalable Sequential Pyramid Networks}
\label{sec:method}

% R3: In figure 1, I would like to suggest the author to add high level
% explanation about designed network architecture.
\begin{figure}[ht]
  \centering
  \includegraphics[width=\textwidth]{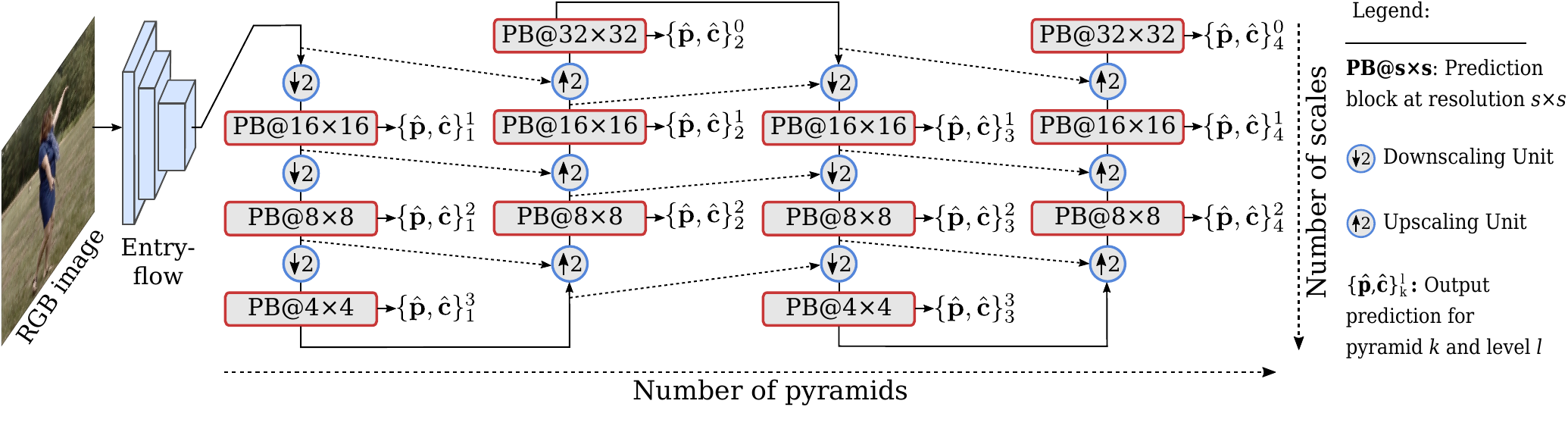}
  \caption{
    Global architecture of SSP-Net.  The entry-flow
    extracts a preliminary feature map from the input image.  These features
    are then fed through a sequence of CNNs composed of
    \textit{prediction blocks} (PB) connected by alternating
    \textit{downscaling} and \textit{upscaling units} (DU and UU).
    Each PB outputs a supervised pose prediction that is refined by further
    blocks and units. See Figures \ref{fig:net-parts} and
    \ref{fig:prediction-block} for the architectural details of DU, UU, and PB.
  }
  \label{fig:netarch}
\end{figure}

\rev{
The proposed network architecture is depicted in Fig.~\ref{fig:netarch}.
The input of our method is an RGB image $\mathbf{I} \in \mathbb{R}^{H\times{W}\times{3}}$
with resolution $H\times{W}$, which is feed to the entry flow network.
The entry-flow produces convolutional features with
resolution $\mathbb{R}^{H/4\times{W/4}\times{384}}$, which are then
fed to a sequence of pyramids. 
The outputs of the network are}
a set of predicted 3D poses,
designated by $\mathbf{p}_k^l \in \mathbb{R}^{{N}\times{3}}$, and
optionally a set of joint confidence scores,
designated by $\mathbf{c}_k^l \in \mathbb{R}^{{N}\times{1}}$, where
$N$ is the number of body joints,
\rev{$k$ is the pyramid index, and $l$ is the level index.
All prediction blocks are supervised during training.
}

\rev{
The motivation for a new architecture design is to provide an explicit
multi-level supervision, enforcing the model to be able to represent the
output independently on the resolution of feature maps. This approach
allows the model to effectively combine low resolution feature maps,
rich in semantic information, with high resolution features,
containing more detailed information.
In order to allow incrementally refined estimations, all predictions
from both low and high resolutions are re-injected into the network.
As a consequence of this densely supervised architecture, the network
can offer early predictions with reduced computational time, or
refined predictions with improved precision.}
The details about the proposed network are presented as follows.

\begin{table}[]
  \centering
  \caption{Entry-flow network.}
  \label{tab:entry-flow}
  \footnotesize
  \begin{tabular}{@{}clccc@{}}
    & Layer                            & Filters & Size/strides & Output \\ \hline
    & Input                            & $3$     &              & $256\times256$ \\ \hline
    & Convolution                      & $64$    & $7\times7/2$ & $128\times128$ \\ \hline
    & Convolution & $64$    & $1\times1$   & \\
    & Convolution & $128$   & $3\times3$   & \\
    & Residual    &         &              & $128\times128$ \\ \hline
    & MaxPooling                       &         & $3\times3/2$ & $64\times64$ \\ \hline
    \multirow{3}{*}{$2\times$\hspace{-3mm}} & Convolution & $128$   & $1\times1$   & \\
    & Convolution & $256$   & $3\times3$   & \\
    & Residual    &         &              & $64\times64$ \\ \hline
    & MaxPooling                       &         & $2\times2/2$ & $32\times32$ \\ \hline
    \multirow{3}{*}{$2\times$\hspace{-3mm}}& Convolution & $192$   & $1\times1$   & \\
    & Convolution & $384$   & $3\times3$   & \\
    & Residual    &         &              & $32\times32$ \\
    \hline
  \end{tabular}
\end{table}

\subsection{Network architecture}

\begin{figure}[htbp]
  \centering
  \begin{subfigure}[b]{0.3\textwidth}
    \centering
    \includegraphics[scale=0.8]{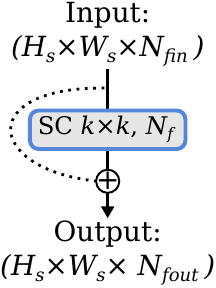}
    \caption{Sep. residual block}
    \label{fig:net-sepconv}
  \end{subfigure}
  \begin{subfigure}[b]{0.3\textwidth}
    \centering
    \includegraphics[scale=0.8]{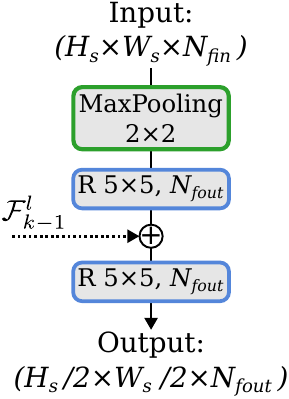}
    \caption{Downscaling unit}
    \label{fig:net-downscaling}
  \end{subfigure}
  \begin{subfigure}[b]{0.3\textwidth}
    \centering
    \includegraphics[scale=0.8]{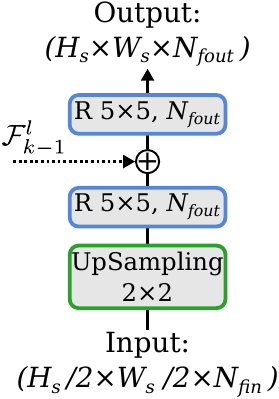}
    \caption{Upscaling unit}
    \label{fig:net-upscaling}
  \end{subfigure}
  \caption{
    Elementary blocks of the proposed network. In (a), the separable residual
    block which is used as the basic building block.
    In (b) and (c), the downscaling unit (DU) and upscaling unit (UU) take as
    secondary input the feature maps $\mathcal{F}_{k-1}^l$ issued from the
    previous pyramid. SC: (depthwise) separable convolution;
    R: separable residual block;
    $H_s\times{W_s}$: features size; $N_{fin/fout}$: number of input/output
    features.
  }
  \label{fig:net-parts}
\end{figure}

The global architecture of the proposed network (Fig.~\ref{fig:netarch})
is essentially composed of a combination of four
modules: \textit{entry-flow}, \textit{downscaling} and \textit{upscaling units},
and \textit{prediction blocks}.
The role of the entry-flow (detailed in Table~\ref{tab:entry-flow}) is to
provide deep convolutional features extraction, which are successively
downscaled and upscaled, respectively by downscaling and upscaling pyramids.
Each pyramid is composed of a sequence of downscaling
or upscaling units (DU or UU, see Fig.~\ref{fig:net-parts}), interleaved with prediction blocks
(PB) at each level. Prediction blocks are indexed by the pyramid index \rev{
$k\in\{1,2,\dots,K\}$, where $K$ is the number of pyramids, and by the
index level $l\in\{0,1,\dots,L\}$, where $L$ is the number of
downscaling or upscaling steps performed,
considering $k=1$ and $l=0$ the CNN features from the entry-flow}.
Note that in this arrangement, an
odd $k$ index corresponds to a downscaling pyramid and an even $k$ index
corresponds to an upscaling pyramid.

The basic building block for the pyramid networks is the separable residual
unit (Fig.~\ref{fig:net-sepconv}), which consists of a depth wise separable
convolution~\cite{Chollet_2017_CVPR} with a residual connection.
Our choice for depth wise separable convolutions is mainly due to its benefits
in efficiency~\cite{HowardZCKWWAA17}.
One important advantage from our approach is the combination of features from
different pyramids and levels. This is performed in both DU/UU, since they
combine features from lower/higher levels, as well as features from previous
pyramids.

\begin{figure}[htbp]
  \centering
  \includegraphics[scale=0.8]{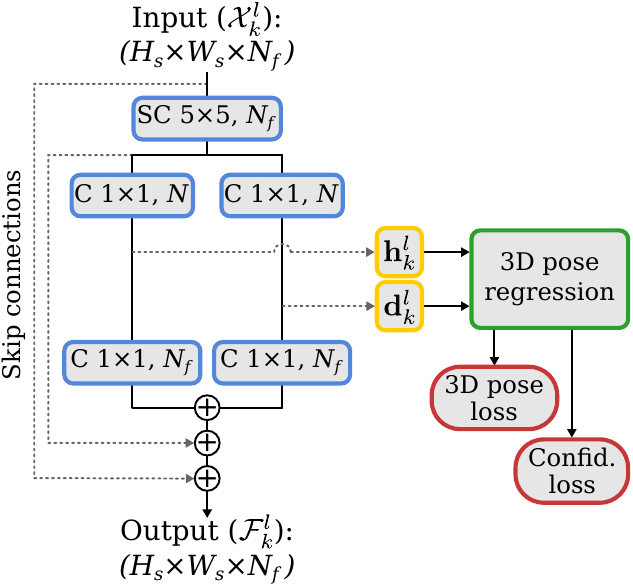}
  \caption{
    Network architecture of the prediction block. Input features $\mathcal{X}_k^l$
    (for pyramid $k$ and level $l$) are used to produce heatmaps
    $\mathbf{h}_k^l$ and depth maps $\mathbf{d}_k^l$, from which 3D pose and
    confidence scores are estimated.  Output features $\mathcal{F}_k^l$ are a
    combination of input features and re-injected predictions.  C: convolution;
    SC: separable convolution; $H_s\times{W_s}$: features size; $N_f$:
    number of features; $N$: number of body joints.
  }
  \label{fig:prediction-block}
\end{figure}

Details of the prediction block (PB) are shown in
Fig.~\ref{fig:prediction-block}. It takes as input a feature map
$\mathcal{X}_k^l$, considering pyramid $k$ and level $l$, and produces a set
of heatmaps $\mathbf{h}_k^l$ and depth maps $\mathbf{d}_k^l$, which are used
for \textit{3D pose regression} (explained in Section~\ref{sec:regression}).
heatmaps and depth maps generation is defined in the following equations:
\begin{equation}
  \mathcal{Y}_k^l = ReLU(BN(SC(\mathcal{X}_k^l))),
\end{equation}
\begin{equation}
  \mathbf{h}_k^l = \mathbf{W}_h^{k,l}\ast{\mathcal{Y}_k^l},
\end{equation}
\begin{equation}
  \mathbf{d}_k^l = \mathbf{W}_d^{k,l}\ast{\mathcal{Y}_k^l},
\end{equation}
where $\mathcal{Y}_k^l$ is an intermediate feature representation,
SC is a separable convolution, $\mathbf{W}_h^{k,l}$
and $\mathbf{W}_d^{k,l}$ are weight matrices with shape
$\mathbb{R}^{N_f\times{N}}$, respectively for heatmaps and depth maps
projection, and $\ast{}$ is the convolution operation.
Additionally, each prediction block also produces a new feature
map $\mathcal{F}_k^l$, which combines the input
features with predicted heatmaps and depth maps, and is used by next blocks
and units for further improvements.
This step is defined in equation~\ref{eq:refine}:
\begin{equation}
  \mathcal{F}_k^l = \mathcal{X}_k^l + \mathcal{Y}_k^l \
    + \mathbf{W}_r^{k,l}\ast{\mathbf{h}_k^l} \
    + \mathbf{W}_s^{k,l}\ast{\mathbf{d}_k^l},
  \label{eq:refine}
\end{equation}
where $\mathbf{W}_r^{k,l}$ and $\mathbf{W}_s^{k,l}$ are called re-injection
matrices.

Differently from the stacked hourglass~\cite{Newell_ECCV_2016,
Pavlakos_2017_CVPR} architectures, where only the higher resolution features
are supervised, we use intermediate supervision at every level of the pyramids.
Adding more supervisions does not significantly increase the computational cost
of our method, since contrarily to the stacked hourglass we do not need
to generate artificial ground truth heatmaps.  On the other hand, with
intermediate supervisions in multiple levels we enforce the robustness of our
method to variations in the scale of feature maps, while efficiently increasing
the receptive field of the global network.
Furthermore, our architecture injects the predictions from these intermediate
supervisions back into the network by merging them with the current features.
This allows the subsequent blocks to perform refining operations instead of
full predictions.

\subsection{3D pose regression approach}
\label{sec:regression}

As discussed in Section~\ref{sec:related-work}, traditional regression
methods use fully connected layers to learn a regression mapping from features
to predictions. However, this approach usually gives sub-optimal solution.
While methods in the state of the art are frequently based on detection,
which requires expensive volumetric heatmap representations,
regression approaches have the advantage of directly providing 3D pose
prediction as joint coordinates without additional post-processing steps.

In our approach, we split the problem as 2D regression and depth estimation,
using two different mappings: heatmaps for $(x, y)$ coordinates and depth maps
for $z$.  For 2D regression, we based our approach on the
\textit{soft-argmax}~\cite{Luvizon_2017_CoRR}, and for depth estimation, we
propose an new attention mechanism guided by 2D joint estimation.  Our method
does not require any parameter and is fully differentiable.  The next sections
explain each part of our approach.

\subsubsection{Soft-argmax for 2D regression}

Let us redefine the softmax operation on a single heatmap $\mathbf{h} \in
\mathbb{R}^{H\times{W}}$ as:
\begin{equation}
  \Phi(\mathbf{h})_{i,j} = \frac{e^{\mathbf{h}_{i,j}}}{%
    \sum_{l=1}^{H}%
    \sum_{c=1}^{W}%
    e^{\mathbf{h}_{l,c}}
    },
  \label{eq:softmax}
\end{equation}
where $\mathbf{h}_{i,j}$ is the value of $\mathbf{h}$ at location
$(i, j)$ and $H\times{W}$ is the heatmap size.
Contrary to the more common cross-channel softmax, we use here a spatial
softmax to ensure that each heatmap is L1 normalized and positive.
Then, we define the soft-argmax as:
\begin{equation}
  \Psi_{d}(\mathbf{h}) = \sum_{i=1}^{H}\sum_{j=1}^{W}%
  \mathbf{W}_{i,j,d}\Phi(\mathbf{h})_{i,j},
  \label{eq:softargmax-math}
\end{equation}
where $\mathit{d}$ is a given component $\mathit{x}$ or $\mathit{y}$, and
$\mathbf{W}$ is a $H\times{W}\times{2}$ weight matrix
for both components $(x, y)$.
The matrix $\mathbf{W}$ can be expressed by its components
$\mathbf{W}_{x}$ and $\mathbf{W}_{y}$, which are 2D discrete normalized ramps,
defined as follows:
\begin{equation}
  \mathbf{W}_{i,j,x} =  \frac{2j-1}{2W}, \mathbf{W}_{i,j,y} = \frac{2i-1}{2H}.
  \label{eq:weights-argmax}
\end{equation}
Finally, given a heatmap $\mathbf{h}$, the regressed location in the image
plane is given by:
\begin{equation}
  \mathbf{\hat{p}}_{img} = (\Psi_{x}(\mathbf{h}), \Psi_{y}(\mathbf{h}))^T.
  \label{eq:regression}
\end{equation}

The soft-argmax operation can be seen as the 2D expectation of the normalized
heatmap, which is a good approximation of the argmax function, considering
that the exponential normalization results in a pointy distribution.

In order to integrate the soft-argmax layer into a deep neural network, we need
its derivative with respect to $\mathbf{h}$:
\begin{equation}
  \frac{\partial \Psi_{d}(\mathbf{h})}{\partial \mathbf{h}_{i,j}} = %
    \mathbf{W}_{i,j,d}%
    \Phi(\mathbf{h})_{i,j}(1-\Phi(\mathbf{h})_{i,j})%
    -\sum_{l=1}^{H}\sum_{c=1}^{W}\mathbf{W}_{l,c,d}%
    \Phi(\mathbf{h})_{i,j}\Phi(\mathbf{h})_{l,c}|_{l\neq{i};c\neq{j}}
  \label{eq:derivative}
\end{equation}
The soft-argmax function can thus be integrated in a trainable framework by
using back propagation and the chain rule on Equation~\ref{eq:derivative}.
Moreover, similarly to what happens on softmax, the gradient is exponentially
increasing for higher values, resulting in very discriminative response
at the joint position.

The soft-argmax layer can be easily implemented in recent frameworks by
concatenating a spatial softmax followed by one non-trainable convolutional
layer with 2 filters of size $H\times{W}$, with fixed parameters according to
Equation~\ref{eq:weights-argmax}.

Unlike traditional argmax, soft-argmax provides sub-pixel accuracy,
allowing good precision even with very low resolution.
Additionally, our approach allows learning very discriminative
heatmaps directly from the $(x, y)$ joint coordinates without explicitly
computing artificial ground truth.
Samples of heatmaps learned by our approach are shown in
Fig.~\ref{fig:samples-heatmaps}.

\subsubsection{Joint based attention for depth estimation}
\label{sec:depth}

For each body joint, we estimate its relative depth $\mathbf{\hat{z}}$ with
respect to the root joint, which is usually designated by the pelvis.
Specifically, we define an
attention mechanism for predicted depth maps based on the appearance
information encoded in heatmaps. Considering one heatmap $\mathbf{h}$ and
the respective depth map $\mathbf{d}$, both with size $\mathbb{R}^{H\times{W}}$,
the estimated relative depth is given by:
\begin{equation}
  \mathbf{\hat{z}} = \frac{%
    \sum_{i=1}^{H}\sum_{j=1}^{W}\mathbf{d}_{i,j} e^{\mathbf{h}_{i,j}}}{%
    \sum_{i=1}^{H}%
    \sum_{j=1}^{W}%
    e^{\mathbf{h}_{i,j}}
    },
  \label{eq:depth}
\end{equation}
which can be interpreted as a selection of relevant regions from $\mathbf{d}$
based on the response from $\mathbf{h}$. In our implementation, values in
depth maps are normalized in the interval $[0, 1]$, corresponding to a range
of depth prediction.

The 3D poses estimated by our approach are composed by the $(x,y)$ coordinates
in pixels (Equation ~\ref{eq:regression}) and by the $z$
coordinate relative to the root joint.
In order to recover the absolute 3D pose in world coordinates, we require the
absolute depth of the root joint and the camera calibration parameters to
convert pixels into millimeters.
We believe that estimating the absolute 3D pose directly in world coordinates
is not the most relevant problem, since the camera calibration can affect such
a prediction drastically. On the other hand, the relative position of joints
with respect to the root is of high relevance, and usually is the only measure
used to compare different methods.
We show in the experiments that absolute depth of the root joint can be
estimated without major impact on accuracy.

\subsubsection{Joint confidence score}
\label{sec:confidence}

\begin{figure}[htbp]
  \centering
  \includegraphics[width=0.18\textwidth]{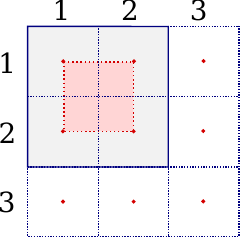}
  \caption{
    Estimation of joint confidence scores. The blue squares represent the pixels
    in the normalized heatmap with its center marked as a red dot.  The red
    square is the region on which a coordinate can be regressed, considering
    responses only on the $2\times2$ window from pixels (1, 1) to (2, 2).
  }
  \label{fig:confidence}
\end{figure}

Additionally to the joint locations, we estimate the joint confidence scores
$\mathbf{\hat{c}}_n$, which corresponds to the probability of the $n^{th}$
joint being visible (or present, even if occluded) in the image.
Given a normalized heatmap, any window with $2\times2$ pixels is enough to
regress a coordinate value with sub-pixel accuracy in a smaller squared region
defined by the centers of the $2\times2$ pixels, as depicted in
Fig.~\ref{fig:confidence}.
Therefore, we apply a summation with a $2\times2$ sliding window on each
normalized heatmap by using a SumPooling with stride 1, and take the maximum
response as the confidence score.
If the normalized heatmap is very pointy, the score is close to 1.
On the other hand, if the normalized heatmap is smooth or has more than one
separated region with high response, the confidence score drops.

Despite giving an additional piece of information, the joint confidence score
does not depend on additional parameters and is computationally negligible,
compared to the cost of the convolutional layers.
Additionally, by supervising this output we can enforce the network to learn
pointy responses for body parts.

%%%%%%%%%%%%%%%%%%%%%%%%%%%%%%%%%%%%%%%%%%%%%%%%%%%%%%%%%%%%%%%%%%%%%%%%%%%%%%
%%%%%%% METHOD
%%%%%%%%%%%%%%%%%%%%%%%%%%%%%%%%%%%%%%%%%%%%%%%%%%%%%%%%%%%%%%%%%%%%%%%%%%%%%%

\section{Experiments}
\label{sec:experiments}

We evaluate the proposed method quantitatively on two challenging datasets for
3D human pose estimation: Human3.6M~\cite{h36m_pami} and
MPI-INF-3DHP~\cite{Mehta_2017_3DV}. We also use the manually annotated
MPII Human Pose dataset (2D only)~\cite{Andriluka_CVPR_2014} to improve the
quality of low level visual features of our network by mixing it with the other
two datasets in a 50\%/50\% ratio on each training batch.
Details about
the 3D human pose datasets used in our experiments are provided as follows.

\subsection{Datasets}

\textbf{Human3.6M}
Human3.6M~\cite{h36m_pami} is a 3D human pose dataset composed of videos with 11
subjects performing 17 different activities, recorded by 4 cameras
simultaneously, resulting in 3.6 million image frames.  For each person, 17
joints are used in our method.
The camera parameters are available, so it is possible to project the 3D joints
to the image plane, as well as the inverse projection from points in the image
plane plus depth back to world coordinates, where the error in computed in
millimeters.
On this dataset, we evaluate our method by measuring the mean per joint
position error (MPJPE), which is a common metric used for this dataset.
We followed the most common evaluation protocol~\cite{Martinez_2017,
Sun_2017_ICCV, Pavlakos_2017_CVPR, Mehta_2017_3DV, Chen_2017_CVPR}
by taking five subjects for training
(S1, S5, S6, S7, S8) and evaluating on two subjects (S9, S11) on one every 64th
frames.
On evaluation, the ground truth and the predicted poses are aligned on the root
joint, and the error is computed on the remaining 16 joints.
As in many similar approaches~\cite{Martinez_2017, Sun_2017_ICCV,
Pavlakos_2017_CVPR}, we use ground truth person bounding boxes for image
cropping and the absolute Z of the root joint to do the inverse projection.
Nonetheless, we demonstrate in the ablation studies (Section \ref{sec:abla})
that errors in the absolute Z of the root joint are much less relevant than
relative joint errors, and we also report our results using estimated absolute
position.

\textbf{MPI-INF-3DHP}
MPI-INF-3DHP~\cite{Mehta_2017_3DV} is, to the best of our knowledge, the most
recent dataset for 3D human pose estimation. It was recorded with a markerless
MoCap system, which allows videos to be recorded in outdoor environment \eg,
TS5 and TS6 from testing. A total of 8 actors were recorded performing 8
activities sets each. The activities involve some complex exercising poses,
which makes this dataset more challenging than Human3.6M.
The authors proposed three evaluation metrics: the mean
per joint position error, in millimeters, the 3D Percentage of Correct
Keypoints (PCK), and the Area Under the Curve (AUC) for different threshold on
PCK. The standard threshold for PCK is 150mm.
Differently from previous work, we use the real 3D poses to compute the error
instead of the normalized 3D poses, since the last one cannot be easily
computed from the image plane.

\subsection{Implementation details}

The proposed network was trained simultaneously on 3D pose regression and on
joint confidence scores.  For pose regression, we used the elastic net loss
function (L1 + L2)~\cite{Zou05regularizationand}:
\begin{equation}
  \mathcal{L}_{\mathbf{p}} = \frac{1}{N_J}%
  \sum_{n=1}^{N_J}%
  \|\mathbf{p}_n - \hat{\mathbf{p}}_n\|_1 +%
  \|\mathbf{p}_n - \hat{\mathbf{p}}_n\|_2^2,
  \label{eq:l1l2-loss}
\end{equation}
where $\mathbf{p}_n$ and $\hat{\mathbf{p}}_n$ are respectively the ground truth
and the predicted $n^{th}$ joint coordinates.
We use directly the joint coordinates normalized to the interval
$[0,1]$, where the top-left image corner corresponds to $(0,0)$, and the
bottom-right image corner corresponds to $(1,1)$.
For the depth ($z$ coordinate), the root joint is assumed to have $z=0.5$,
and a range of 2 meters is used to represent the remaining joints, which means
that $z=0$ corresponds to a depth of $-1$ meter with respect to the root.

For the joint confidence scores, we use the binary cross entropy loss
function:
\begin{equation}
  \mathcal{L}_{\mathbf{c}} = \frac{1}{N_J}%
  \sum_{n=1}^{N_J}%
  [(\mathbf{c}_n-1)~log~(1-\hat{\mathbf{c}}_n)%
  -\mathbf{c}_n~log~\hat{\mathbf{c}}_n],
  \label{eq:loss-prob}
\end{equation}
where $\mathbf{c}_n$ and $\hat{\mathbf{c}}_n$ are respectively the ground
truth and the predicted confidence scores. We use $\mathbf{c}_n = 1$ if the
$n^{th}$ joint is present in the image and $\mathbf{c}_n = 0$ otherwise.

The network architecture used in our experiments is implemented according to
Fig.~\ref{fig:netarch} and is composed of 8 pyramids, divided as 4 downscaling
and 4 upscaling pyramids, each one with 4 scales ($K = 8$ and $L = 3$).
We optimize the network using back propagation and RMSprop
with batches of 24 images and initial learning rate of 0.001, which is divided
by 10 when validation score plateaus.
%
%The full training of our network takes about two days on a GeForce GTX 1080 Ti
%with 11GB of memory.
%
We used standard data augmentation on all datasets, including: random
rotations ($\pm30^{\circ}$), random bounding box rescaling with a factor from
$0.7$ to $1.3$, and random brightness gain on color channels from $0.9$ to
$1.1$.

\subsection{Results on 3D pose estimation}

\begin{figure}[hbpt!]
  \centering
    \includegraphics[width=2.3cm]{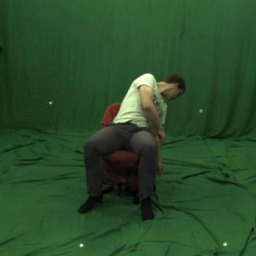}\hspace{0.3mm}
    \includegraphics[width=2.3cm]{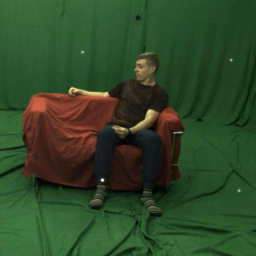}\hspace{0.3mm}
    \includegraphics[width=2.3cm]{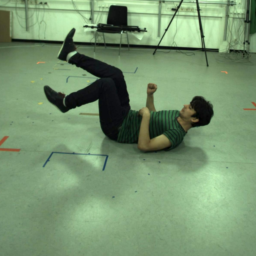}\hspace{0.3mm}
    \includegraphics[width=2.3cm]{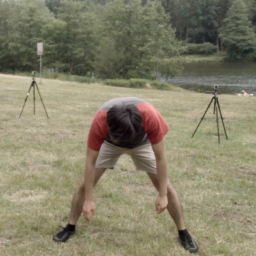}\hspace{0.3mm}
    \includegraphics[width=2.3cm]{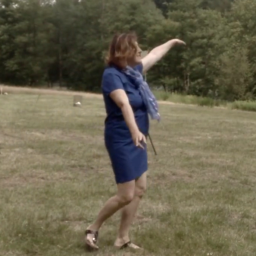}\hspace{0.3mm}\\
    \includegraphics[width=2.3cm]{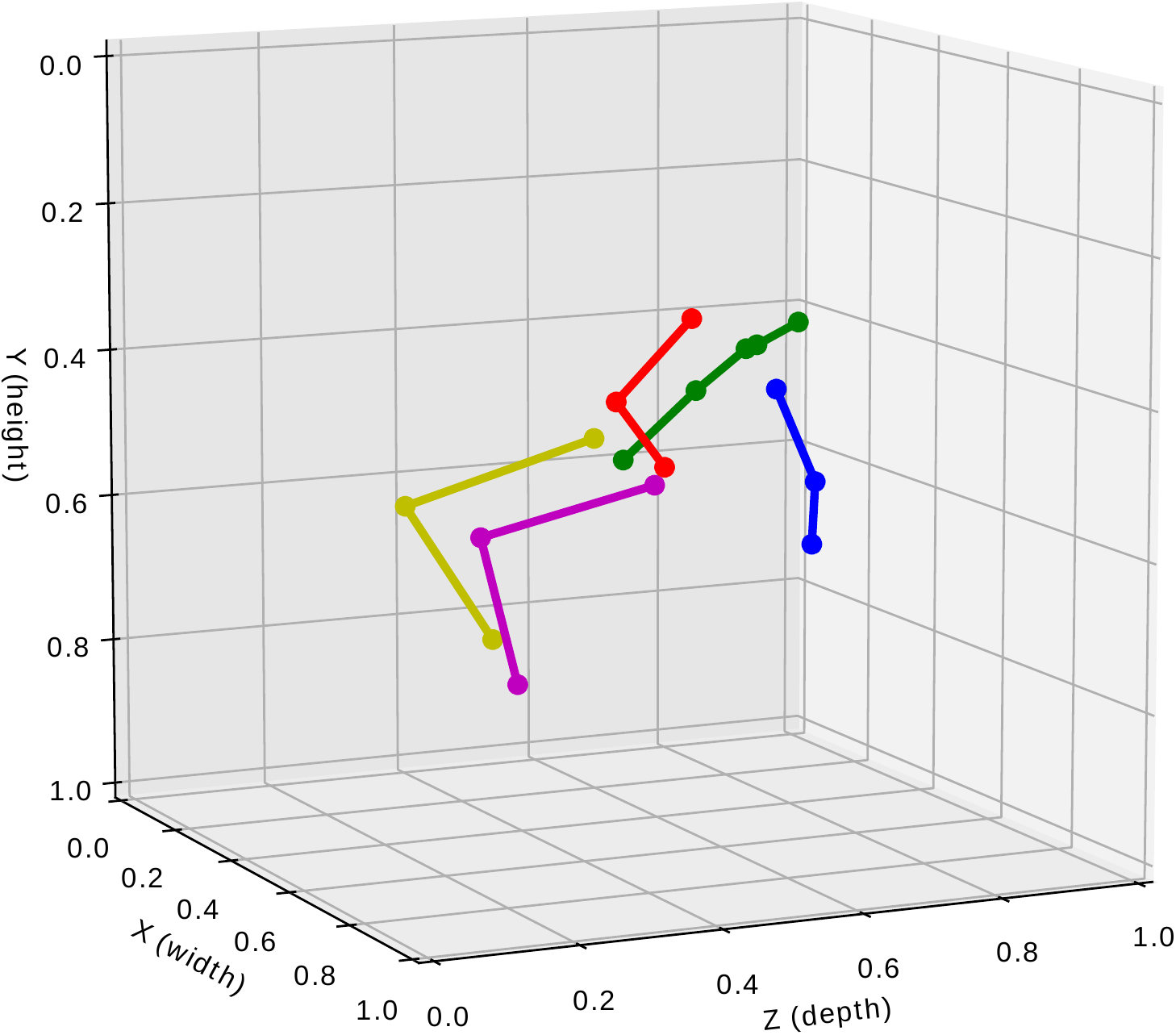}\vspace{0.3mm}
    \includegraphics[width=2.3cm]{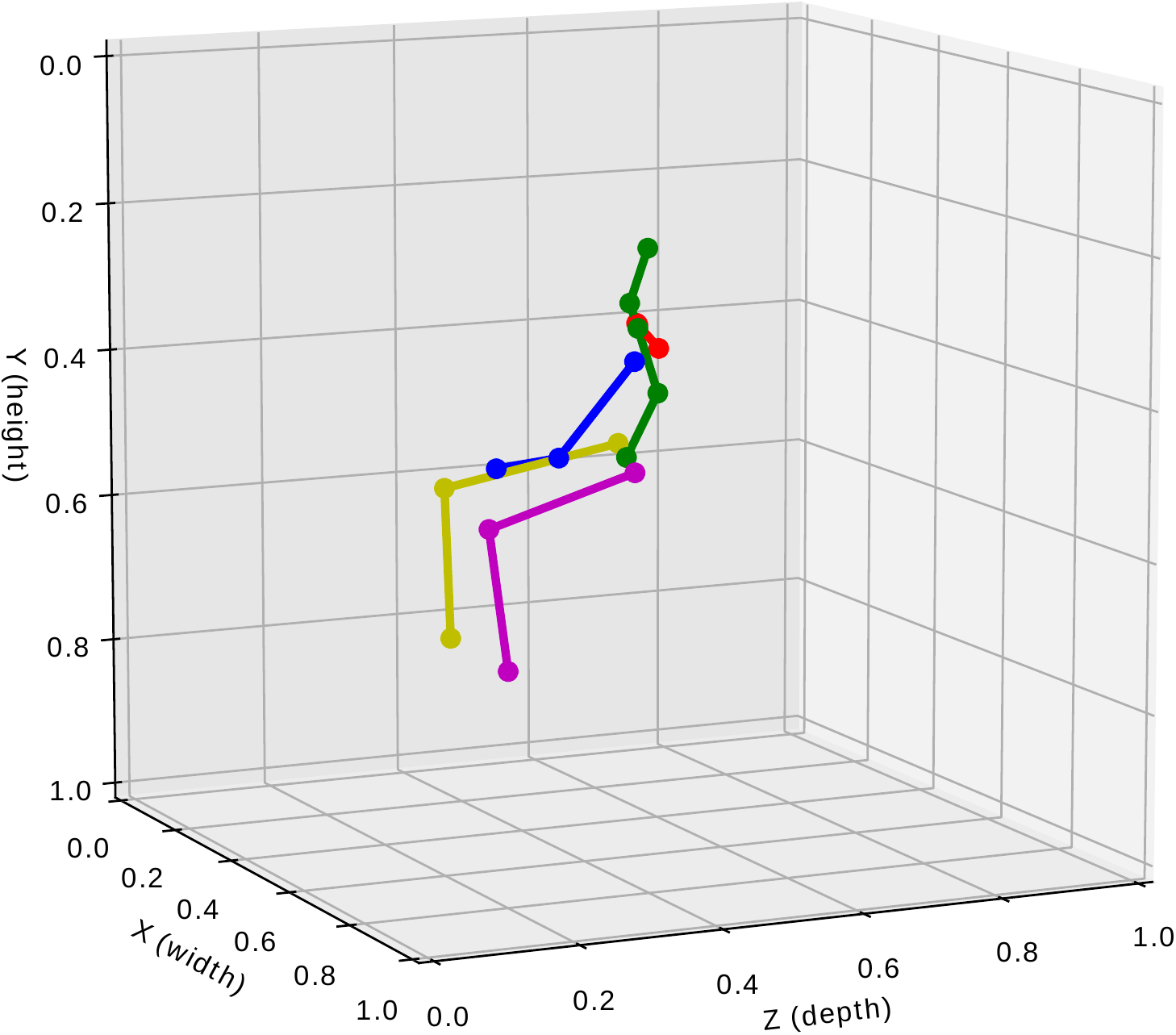}\vspace{0.3mm}
    \includegraphics[width=2.3cm]{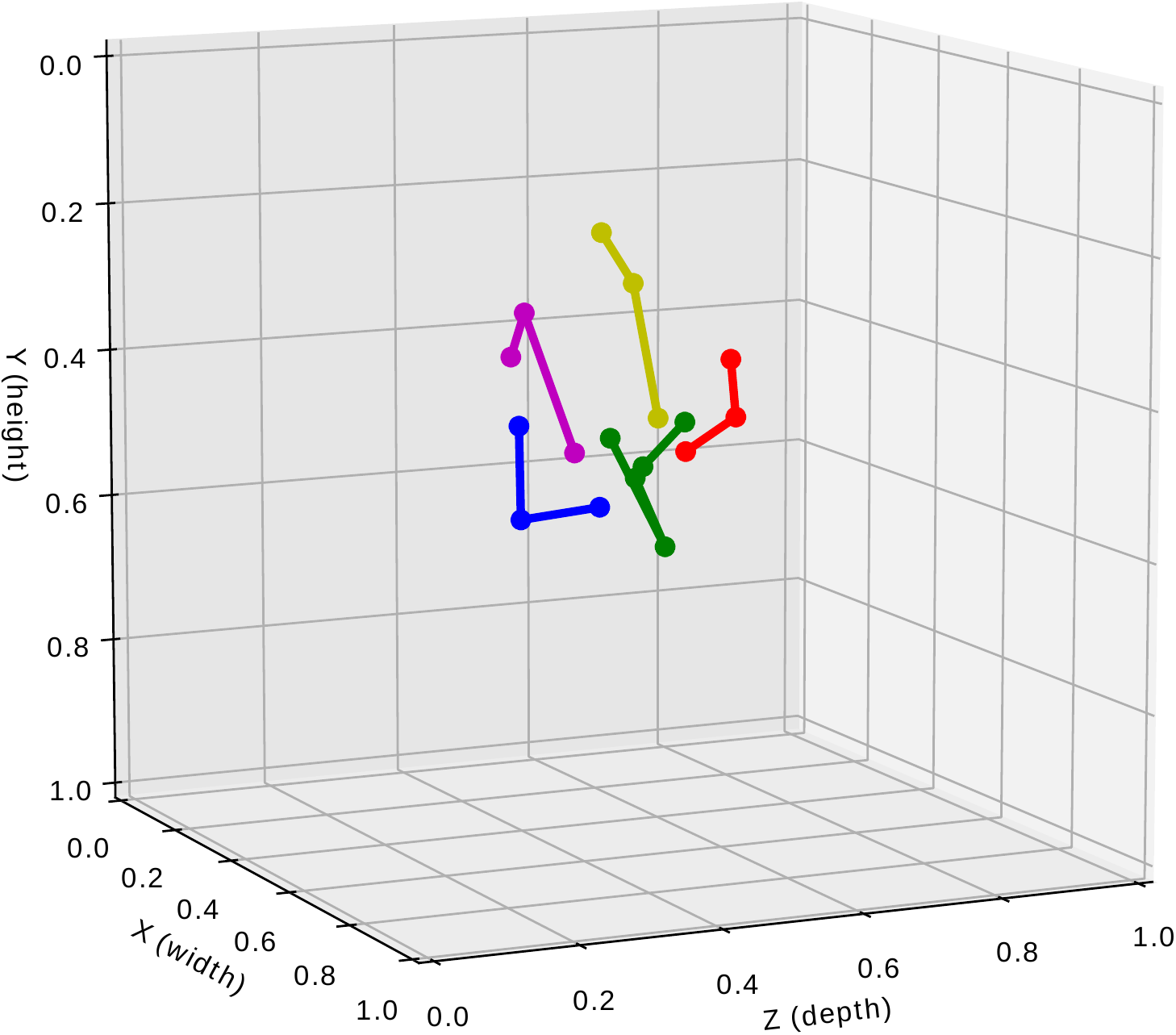}\vspace{0.3mm}
    \includegraphics[width=2.3cm]{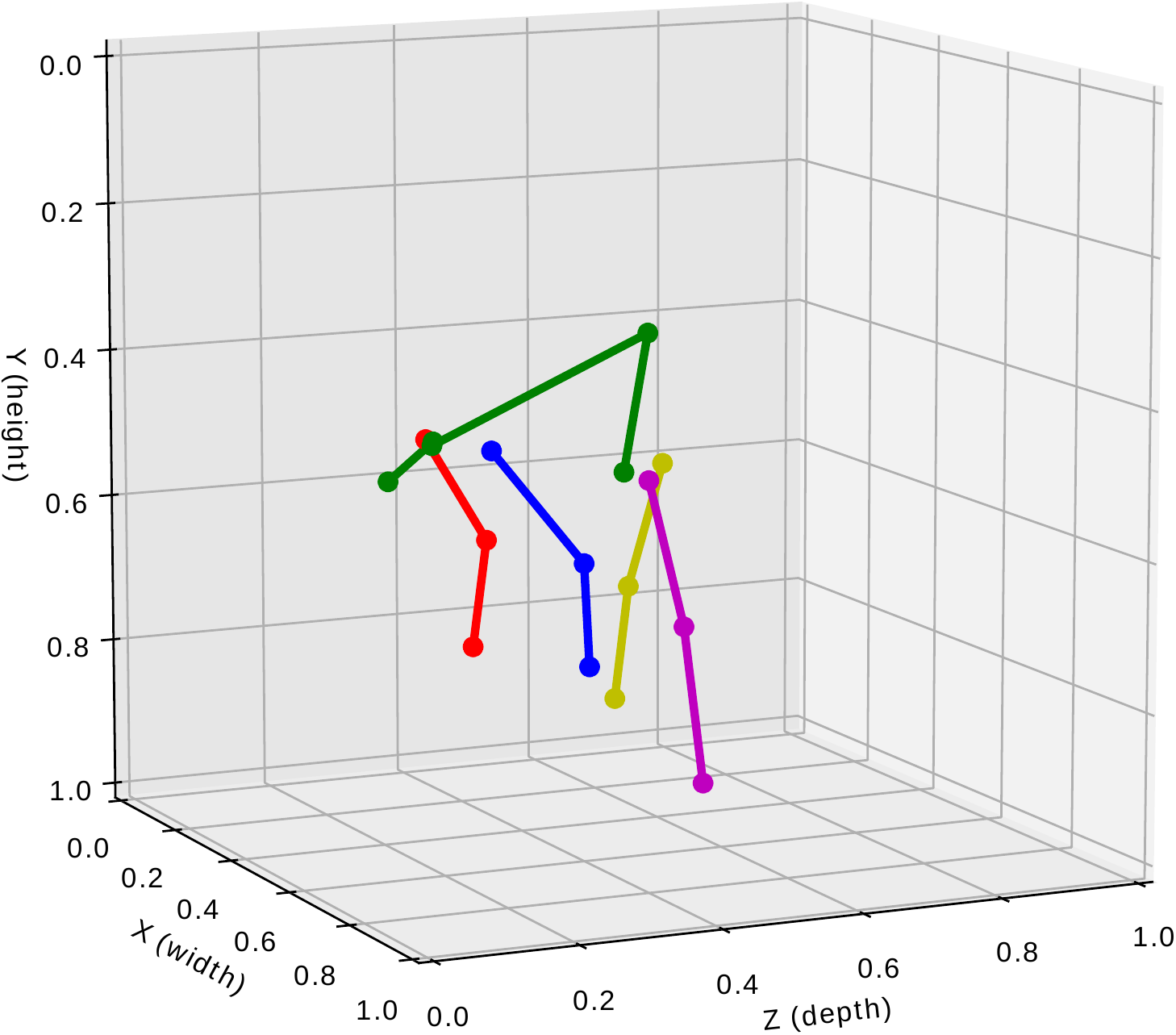}\vspace{0.3mm}
    \includegraphics[width=2.3cm]{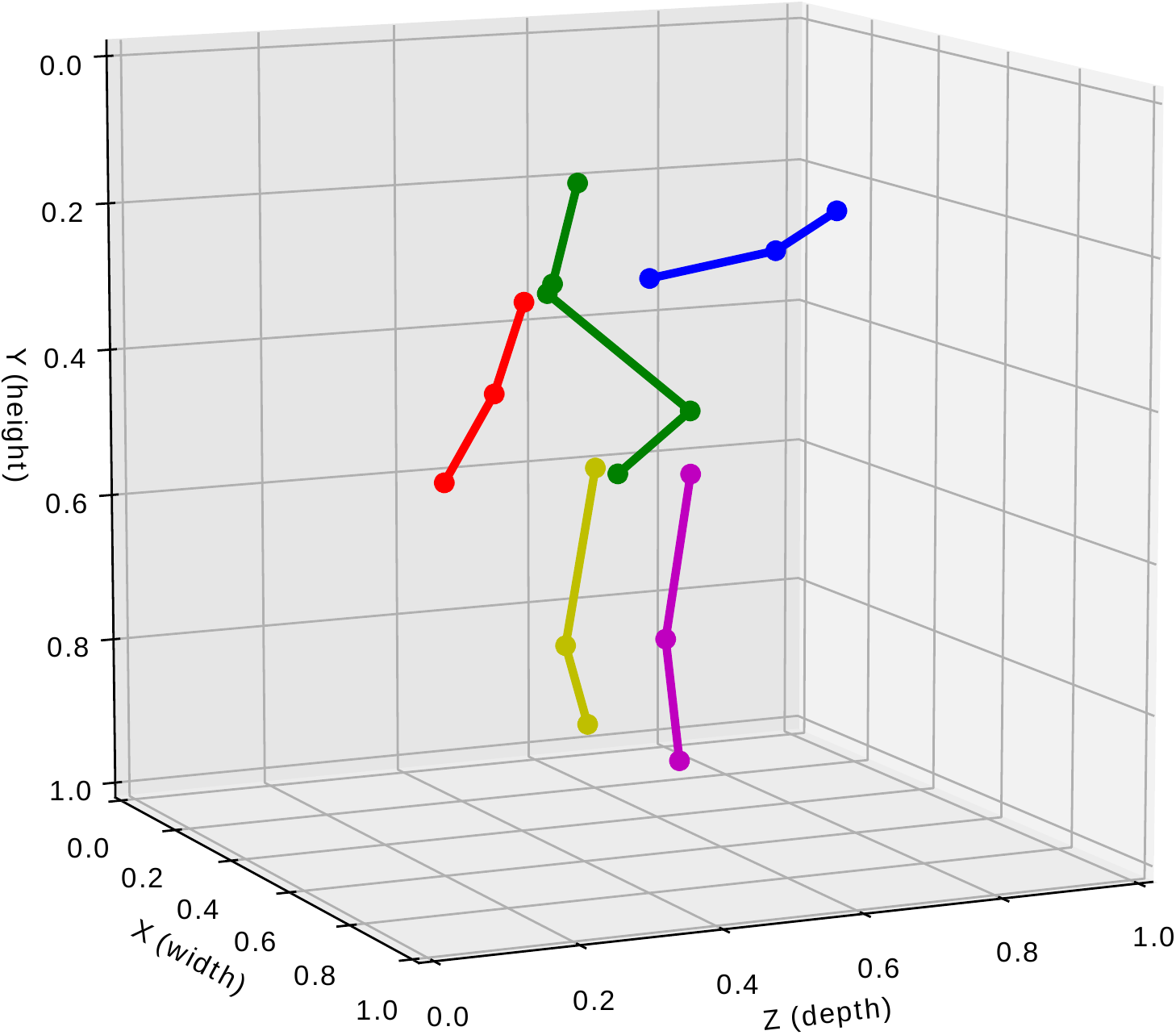}\vspace{0.3mm}
  \caption{
    Input image samples (top), and their respective predicted 3D poses (bottom) for the MPI-INF-3DHP dataset, including two outdoor scenes (from testing).
  }
  \label{fig:samples-3dposes}
\end{figure}

\rev{Figure~\ref{fig:samples-3dposes} shows some qualitative results
of our method for 3D pose estimation, including challenging poses and some outdoor scenes. A quantitative evaluation is presented as follows.}

\begin{table}[ht]
  \centering
  \caption{Comparison results with previous work on Human3.6M using the MPJPE
  (millimeters errors) evaluation on reconstructed poses. AZ: using the
  absolute $z$ of the root joint. MC: multi-crop, using 5 different bounding
  boxes with horizontal flip.}
  \label{tab:result-h36m}
  %\small
  \footnotesize
  \begin{tabular}{@{}l|cccccccc@{}}
    \hline
    Methods & Dir. & Disc. & Eat & Greet & Phone & Posing & Purch. & Sit \\ \hline
        \hline

    Pavlakos \etal \cite{Pavlakos_2017_CVPR}    & 67.4 & 71.9 & 66.7 & 69.1 & 71.9 & 65.0 & 68.3 & 83.7 \\
    Mehta \etal \cite{Mehta_2017_3DV}           & 52.5 & 63.8 & 55.4 & 62.3 & 71.8 & 52.6 & 72.2 & 86.2 \\
    Martinez \etal \cite{Martinez_2017}         & 51.8 & 56.2 & 58.1 & 59.0 & 69.5 & 55.2 & 58.1 & 74.0 \\
    Luo \etal \cite{luo2018orinet}$^\ast{}$     & 49.2 & 57.5 & 53.9 & 55.4 & 62.2 & 52.1 & 60.9 & 73.8 \\
    Sun \etal \cite{Sun_2017_ICCV}              & 52.8 & 54.8 & 54.2 & 54.3 & 61.8 & 53.1 & 53.6 & 71.7 \\
    Luvizon \etal \cite{Luvizon_2018_CVPR}      & 49.2 & 51.6 & \textbf{47.6} & 50.5 & 51.8 & 48.5 & 51.7 & 61.5 \\
    Sun \etal \cite{Sun_2018_ECCV}              & -- & -- & -- & -- & -- & -- & -- & -- \\ \hline
    \textbf{Ours \@ 120 FPS}  & \textbf{46.9} & \textbf{50.9} & 49.9 & \textbf{47.5} & 51.9 & \textbf{46.2} & \textbf{49.2} & 61.7 \\
    \textbf{Ours} +AZ  & 46.1 & 50.2 & 50.2 & 47.5 & 52.0 & 45.9 & 48.5 & 62.3 \\
    \textbf{Ours} +AZ+MC & \textbf{45.1} & \textbf{49.1} & 49.0 & \textbf{46.5} & \textbf{50.6} & \textbf{44.8} & \textbf{47.7} & \textbf{60.6} \\
    \hline
    Methods & SitD. & Smoke & Photo & Wait & Walk & WalkD. & WalkP. & \multicolumn{1}{|c@{}}{Avg} \\ \hline
        \hline
        
    Pavlakos \etal \cite{Pavlakos_2017_CVPR}    & 96.5 & 71.4 & 76.9 & 65.8 & 59.1 & 74.9 & 63.2 & \multicolumn{1}{|c@{}}{71.9} \\
    Mehta \etal \cite{Mehta_2017_3DV}           & 120.0& 66.0 & 79.8 & 63.9 & 48.9 & 76.8 & 53.7 & \multicolumn{1}{|c@{}}{68.6} \\
    Martinez \etal \cite{Martinez_2017}         & 94.6 & 62.3 & 78.4 & 59.1 & 49.5 & 65.1 & 52.4 & \multicolumn{1}{|c@{}}{62.9} \\
    Luo \etal \cite{luo2018orinet}\textsuperscript{$\star$}     & 96.5 & 60.4 & 73.9 & 55.6 & 46.6 & 69.5 & 52.4 & \multicolumn{1}{|c@{61.3}}{} \\
    Sun \etal \cite{Sun_2017_ICCV}              & 86.7 & 61.5 & 67.2 & 53.4 & 47.1 & 61.6 & 53.4 & \multicolumn{1}{|c@{}}{59.1} \\
    Luvizon \etal \cite{Luvizon_2018_CVPR}      & 70.9 & 53.7 & 60.3 & 48.9 & 44.4 & 57.9 & 48.9 & \multicolumn{1}{|c@{}}{53.2} \\
    Sun \etal \cite{Sun_2018_ECCV}              & -- & -- & -- & -- & -- & -- & -- & \multicolumn{1}{|c@{}}{\textbf{49.6}} \\ \hline
    \textbf{Ours \@ 120 FPS}      & \textbf{66.5} & \textbf{53.4} & \textbf{55.2} & \textbf{45.5} & \textbf{42.1} & \textbf{55.6} & \textbf{45.9} & \multicolumn{1}{|c@{}}{\textbf{51.6}} \\
    \textbf{Ours} +AZ & 66.8 & 53.4 & 54.7 & 45.2 &  41.9 & 54.7 & 45.5 & \multicolumn{1}{|c@{}}{51.4} \\
    \textbf{Ours} +AZ+MC  & \textbf{65.4} & \textbf{52.0} & \textbf{52.8} & \textbf{44.2} & \textbf{40.6} & \textbf{54.1} & \textbf{44.4} & \multicolumn{1}{|c@{}}{\textbf{50.2}} \\\hline
  \end{tabular}\\
  \small{\textsuperscript{$\star$} Results using ground truth limb lengths.}
\end{table}

\subsubsection{Human3.6M} Table~\ref{tab:result-h36m} shows our results compared
to recent methods, where we achieve \textbf{50.2 mm} average MPJPE considering
multi-crop and \textbf{51.6 mm} single-crop at 120 frames per second (FPS). Our approach
achieves results comparable to the state-of-the-art overall, and improves individual
activities up to 12.4\% on ``Photo'' and 7.7\% on ``Sit down'', which is the
most challenging case. In general, our method improves state-of-the-art on
individual activities even on single-crop at full speed, running on a desktop
GeForce GTX 1080Ti GPU, which is, to the best of our knowledge, better than any
previous method.
Additionally, with the proposed architecture, our approach can be even faster
with a small decrease in performance, as shown in the ablation study.

We also evaluate our method using the estimated Z (depth) of root joints, which
corresponds to our results when nothing else is specified. For doing that, we
use a MLP with three layers and 128-128-256 neurons, which takes as input the
image bounding box normalized coordinates (2D only) and a vector of visual
features ($\mathcal{F}_1^3$), and outputs the estimated absolute Z of the root joint.

\subsubsection{MPI-INF-3DHP}
Our results on this dataset are presented in
Table~\ref{tab:result-mpii}. \rev{We reached a comparable result to Luo \etal~\cite{luo2018orinet},
Improving their result on the average PCK by 1.4\%, while producing inferences much
faster (120 FPS on a GTX 1080 Ti \textit{vs} 20 FPS from \cite{luo2018orinet}
on a Titan XP). Furthermore, we are the only method to not use the \textit{universal} normalized poses from this dataset, since our method requires the full pose in its original coordinates to allows camera projection.
}

\begin{table}[ht]
  \centering
  \caption{Comparison results with previous work on MPI-INF-3DHP using the PCK
  and AUC metrics (higher is better) and the MPJPE metric (lower is better), on
  reconstructed poses.  AZ: using the absolute $z$ of the root joint.}
  \label{tab:result-mpii}
  \scriptsize
  \begin{tabular}{@{}p{22mm}| p{5mm} p{5mm} p{5mm} p{6mm} p{6mm} p{5mm} p{5mm} p{5mm} p{5mm} p{8mm} @{}}
    \hline
    Methods & \begin{tabular}[c]{@{}l@{}}Std.\\Walk\end{tabular} & Exer. & \begin{tabular}[c]{@{}l@{}}Sit\\Chair\end{tabular} & \begin{tabular}[c]{@{}l@{}}Croush\\Reach\end{tabular} & \begin{tabular}[c]{@{}l@{}}OnThe\\Floor\end{tabular} & Sport & Misc. & \multicolumn{3}{|c}{Avg} \\ \hline
      & PCK & PCK & PCK & PCK & PCK & PCK & PCK & \multicolumn{1}{|c}{PCK} & AUC & \tiny{MPJPE} \\ \hline
        \hline
    Zhou \etal \cite{Zhou_2017_ICCV}\textsuperscript{$\star{}$}  & - & - & - & - & - & - & - & \multicolumn{1}{|c}{69.2} & 32.5 & - \\
    Mehta \etal \cite{Mehta_2017_3DV}\textsuperscript{$\star{}$}      & 86.6 & 75.3 & 74.8 & 73.7 & 52.2 & 82.1 & 77.5 & \multicolumn{1}{|c}{75.7} & 39.3 & 117.6 \\
    Mehta \etal \cite{VNect_SIGGRAPH2017}\textsuperscript{$\star{}$}  & 87.7 & 77.4 & 74.7 & 72.9 & 51.3 & 83.3 & 80.1 & \multicolumn{1}{|c}{76.6} & 40.4 & 124.7 \\
    Luo \etal \cite{luo2018orinet}\textsuperscript{$\star{}$} & \textbf{90.4} & 79.1 & \textbf{88.5} & \textbf{81.6} & 66.3 & \textbf{91.9} & \textbf{92.2} & \multicolumn{1}{|c}{81.8} & \textbf{45.2} & \textbf{89.4} \\
        \hline
    \textbf{Ours} +AZ & 87.1 & 85.4 & 85.9 & \textbf{81.6} & \textbf{68.5} & 88.2 & 83.0 & \multicolumn{1}{|c}{\textbf{83.2}} & 44.3 & 96.8 \\ \hline
  \end{tabular}\\
  \small{\textsuperscript{$\star$} Results using the \textit{universal} (normalized) ground truth poses.}
\end{table}

\subsection{Ablation study}
\label{sec:abla}

In this section we provide some additional experiments that show the behaviour
of our method with respect to the proposed network architecture.

\begin{figure}[htbp]
  \centering
  \begin{subfigure}[b]{0.31\textwidth}
    \includegraphics[width=\textwidth]{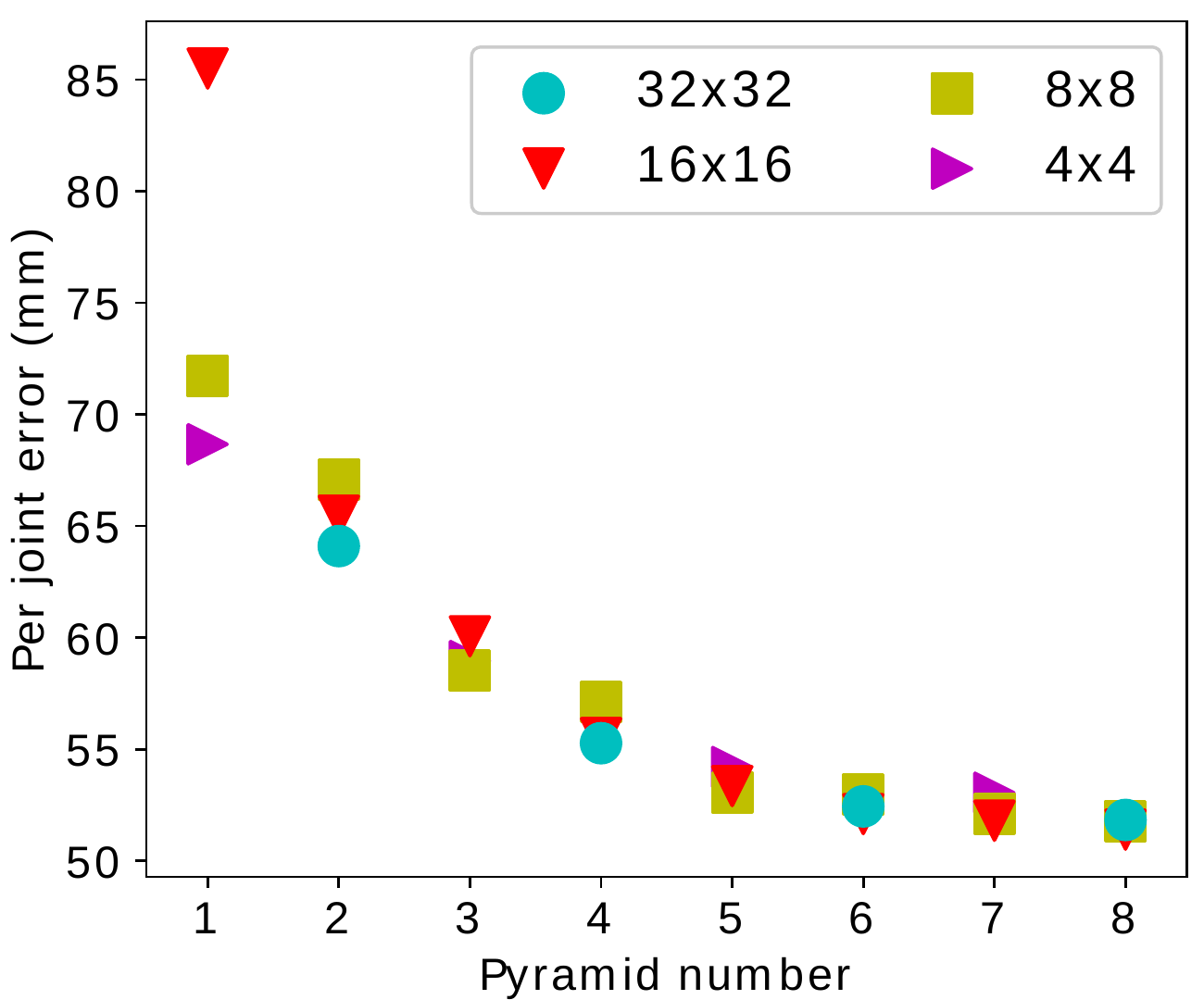}
    \caption{}
    \label{fig:abl01}
  \end{subfigure}
  \begin{subfigure}[b]{0.33\textwidth}
    \includegraphics[width=\textwidth]{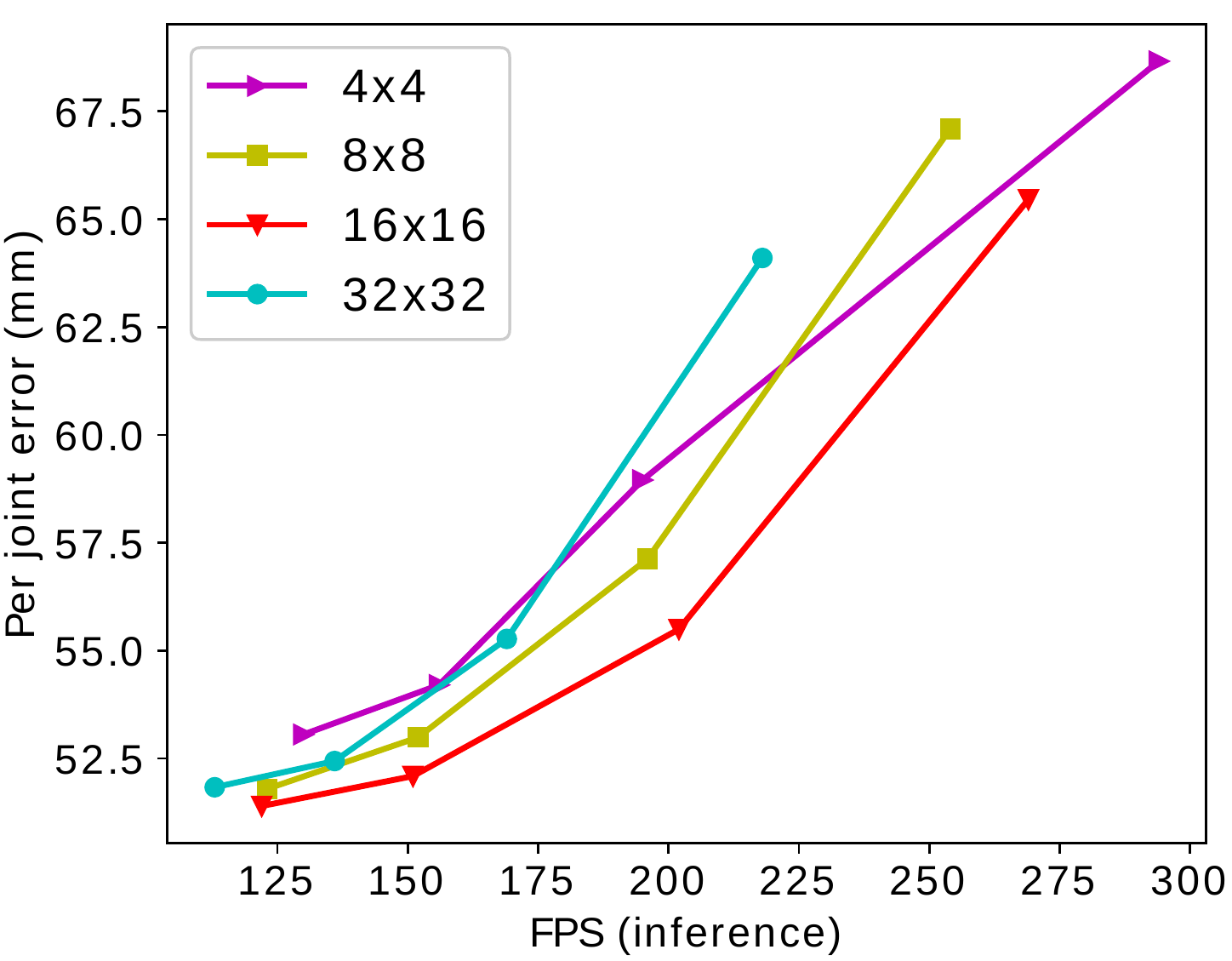}
    \caption{}
    \label{fig:abl02}
  \end{subfigure}
  \begin{subfigure}[b]{0.31\textwidth}
    \includegraphics[width=\textwidth]{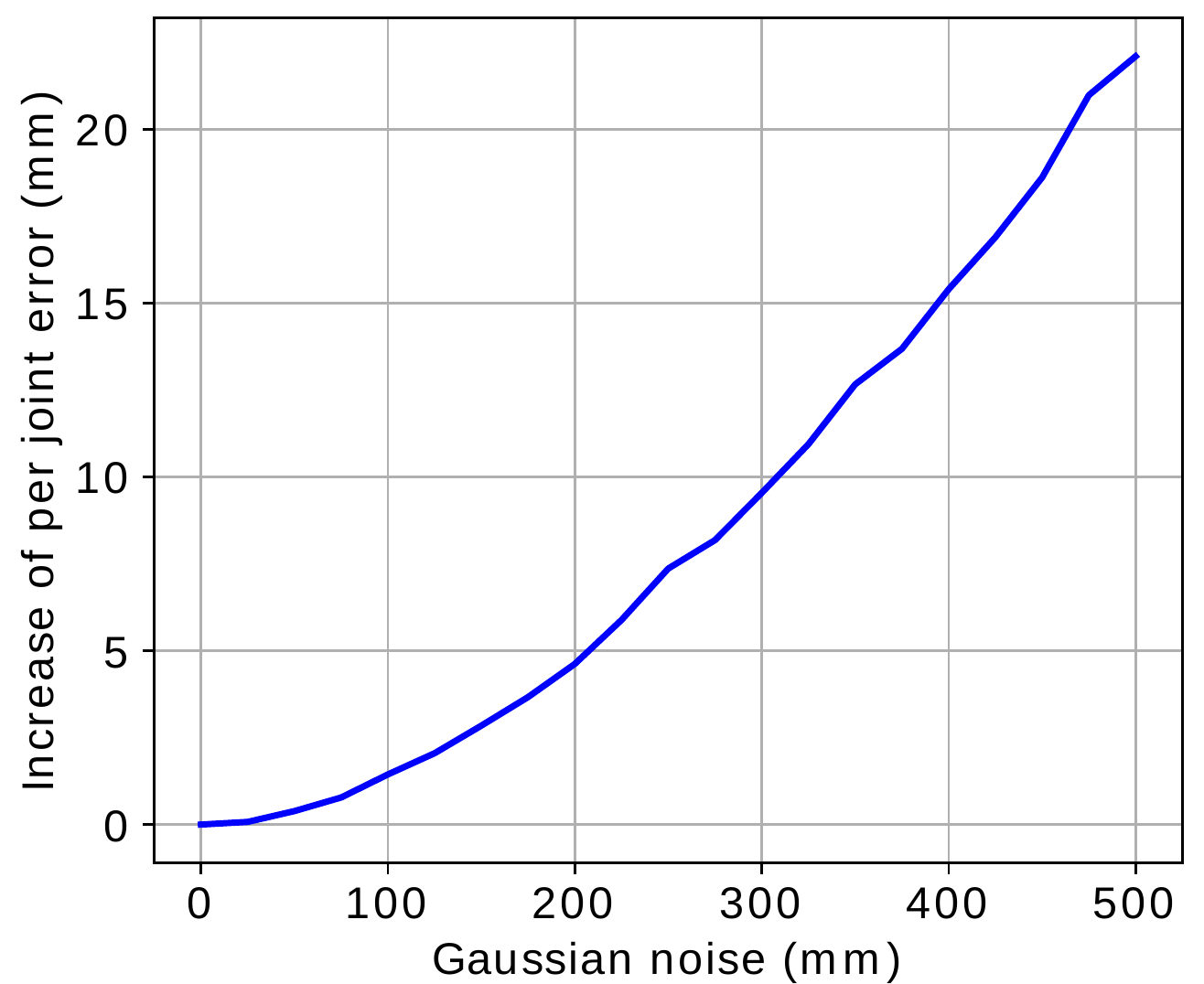}
    \caption{}
    \label{fig:abl03}
  \end{subfigure}
  \caption{
    Ablation study of our method. In (a), we shown the error performed by each
    intermediate supervision. The trade off between precision (related to the
    number of pyramids) and speed is shown in (b), for all the pyramid levels.
    In (c) we present the increase in reconstruction error with respect to a
    Gaussian noise injected on absolute root joint position.
    %(a) and (b) are
    %performed on Human3.6M and (c) is performed on MPI-INF-3DHP.
  }
  \label{fig:ablationstudy}
\end{figure}

In Fig.~\ref{fig:abl01}, we consider each intermediate supervision of the
network as a valid output and we show the improvement on accuracy
(error decreasing) with respect to the number of pyramids in the network.
Additionally, the error with respect to each pyramid scale is also shown. We
can clearly see that all the scales are improved by the sequence of pyramids,
in such a way that in the last pyramid all scales present very similar error.
This evolution can be better seen in Table~\ref{tab:all}, where the error of all
intermediate predictions are shown.
Note that the precision of our regression method is invariant to the scale of
the feature maps, since we reached excellent results with heatmaps of
$4\times4$ pixels. The same is not true for detection based approach, like in
\cite{Pavlakos_2017_CVPR}, since in their method the predictions are quantized
by the argmax function. The error introduced by this quantization can be observed
in Table~\ref{tab:result-quantization}, where we compare our regression approach with
ground truth volumetric heatmaps and argmax.

\begin{table}[ht]
  \centering
 \caption{Results on Human3.6M (millimeters error), comparing predictions
  using ground truth heatmaps and argmax \textit{vs.} our regression approach.}
  \label{tab:result-quantization}
%% Results using GT heatmaps + argmax:
% Average reconstruction error (4x4): 233.878 mm, 97.800 pix
% Average reconstruction error (8x8): 128.610 mm, 47.394 pix
% Average reconstruction error (16x16): 59.929 mm, 23.639 pix
% Average reconstruction error (32x32): 30.984 mm, 12.239 pix
% Average reconstruction error (64x64): 15.735 mm, 6.235 pix
  \footnotesize
  \begin{tabular}{l|c|c|c|c}
    \hline
    Method / resolution & $s=4$ & $s=8$ & $s=16$ & $s=32$ \\ \hline
    \hline
    Volumetric GT heatmaps ($s\times s \times s$) + argmax  & 233.9 & 128.6 & 59.9  & 31.0 \\ \hline
    \textbf{Our regression approach} (soft-argmax) & 53.0 & 51.8 & 51.4 & 51.6 \\ \hline
  \end{tabular}
\end{table}

\begin{figure}[hbpt!]
  \centering
  \begin{subfigure}[b]{0.18\textwidth}
    \includegraphics[height=2.1cm]{000950_img.png}\vspace{0.2mm}\\
    \includegraphics[height=2.1cm]{001680_img.png}\vspace{0.2mm}\\
    \includegraphics[height=2.1cm]{002680_img.png}\vspace{0.2mm}\\
    \caption{}
  \end{subfigure}\hspace{2mm}
  \begin{subfigure}[b]{0.18\textwidth}
    \includegraphics[height=2.1cm]{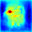}\vspace{0.2mm}\\
    \includegraphics[height=2.1cm]{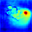}\vspace{0.2mm}\\
    \includegraphics[height=2.1cm]{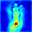}\vspace{0.2mm}\\
    \caption{}
  \end{subfigure}
  \begin{subfigure}[b]{0.18\textwidth}
    \includegraphics[height=2.1cm]{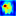}\vspace{0.2mm}\\
    \includegraphics[height=2.1cm]{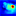}\vspace{0.2mm}\\
    \includegraphics[height=2.1cm]{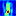}\vspace{0.2mm}\\
    \caption{}
  \end{subfigure}
  \begin{subfigure}[b]{0.18\textwidth}
    \includegraphics[height=2.1cm]{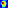}\vspace{0.2mm}\\
    \includegraphics[height=2.1cm]{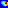}\vspace{0.2mm}\\
    \includegraphics[height=2.1cm]{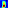}\vspace{0.2mm}\\
    \caption{}
  \end{subfigure}
  \begin{subfigure}[b]{0.18\textwidth}
    \includegraphics[height=2.1cm]{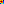}\\
    \includegraphics[height=2.1cm]{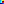}\\
    \includegraphics[height=2.1cm]{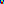}\\
    \caption{}
  \end{subfigure}
  \caption{
    Input image samples (a), and their respective heatmaps indirectly learned for selected joints
    at different pyramid levels (b, c, d, e).
  }
  \label{fig:samples-heatmaps}
\end{figure}

One important characteristic of our network is that it offers an excellent
trade off between performance and speed. In Fig.~\ref{fig:abl02} we show the
per joint error for four pyramids with their respective scales compared to the
inference speed. Note that we are able to reach 55.5 millimeters error, which
is still a good result on Human3.6M, at a very fast inference rate of 200 FPS.
\rev{Additionally, in Fig~\ref{fig:samples-heatmaps} we show the our approach is able to learn very low resolution heatmap representations, while still achieving competitive results.}

Finally, we demonstrate on Fig.~\ref{fig:abl03} the influence of a bad prediction
of the absolute root depth by adding a Gaussian noise on the ground truth
reference. By adding a noise of 100 millimeters (about the same magnitude of
the precision of our method on MPI-INF-3DHP), we have an increase in error
inferior to 2 millimeters.  This clearly reinforces our idea that the error on
relative joint positions is much more relevant than the absolute offset of the
root joint.

\begin{table}[]
  \centering
  \caption{Mean per joint position error (MPJPE) in millimeters for all
  intermediate supervisions of the SSP-Net on Human3.6M.  Odd pyramid
  numbers correspond to Downscaling Pyramids, and even numbers correspond to
  Upscaling Pyramids.}
  \label{tab:all}
  \small
  \begin{tabular}{cc|cccccccc}
    \hline
    Scale & Features res. & \multicolumn{8}{c}{\begin{tabular}[c]{@{}c@{}}Pyramid number / MPJPE\end{tabular}} \\ \hline
            &              & ~~1~~ & ~~2~~ & ~~3~~ & ~~4~~ & ~~5~~ & ~~6~~ & ~~7~~ & ~~8~~       \\ \hline
    \hline
    $L^0$   & $32\times32$ & -       & 64.1    & -       & 55.3    & -       & 52.4    & -       & 51.6   \\ \hline
    $L^1$   & $16\times16$ & 85.5    & 65.5    & 60.1    & 55.5    & 55.3    & 52.1    & 51.8    & 51.4    \\ \hline
    $L^2$   & $8\times8$   & 71.7    & 67.1    & 58.5    & 57.1    & 53.1    & 53.0    & 52.1    & 51.8    \\ \hline
    $L^3$   & $4\times4$   & 68.7    & -       & 58.9    & -       & 54.2    & -       & 53.0    & -       \\ \hline
\end{tabular}
\end{table}

\section{Conclusion}
\label{sec:conclusion}

In this work, we have presented a new regression method and a new scalable
network architecture for 3D human pose estimation from still RGB images.
The method is based on the proposed Scalable Sequential Pyramid Networks, which
is a highly scalable network that can be very precise at a small computational
cost and extremely fast with a small decrease in accuracy, with a single
training procedure.
The proposed parameter free regression approach is invariant to the resolution
of feature maps thanks to the soft-argmax operation, while performing
state-of-the-art scores on important benchmarks for 3D pose estimation.
Additionally, we provided some intuitions about the behaviour of our method
in our ablation study, which demonstrates its effectiveness, specially for
efficient predictions.

\section{Acknowledgements}

This work was partially founded by CNPq (Brazil) - Grant 233342/2014-1.

%% If you have bibdatabase file and want bibtex to generate the
%% bibitems, please use
%%
%%  \bibliographystyle{elsarticle-num} 
%%  \bibliography{<your bibdatabase>}

%% else use the following coding to input the bibitems directly in the
%% TeX file.

%TODO Add the acknowledgement latter

\bibliographystyle{elsarticle-num}
\bibliography{references}

\begin{thebibliography}{10}
\expandafter\ifx\csname url\endcsname\relax
  \def\url#1{\texttt{#1}}\fi
\expandafter\ifx\csname urlprefix\endcsname\relax\def\urlprefix{URL }\fi
\expandafter\ifx\csname href\endcsname\relax
  \def\href#1#2{#2} \def\path#1{#1}\fi

\bibitem{Luvizon_2018_CVPR}
D.~C. Luvizon, D.~Picard, H.~Tabia, 2d/3d pose estimation and action
  recognition using multitask deep learning, in: The IEEE Conference on
  Computer Vision and Pattern Recognition (CVPR), 2018.

\bibitem{PISHCHULIN2017276}
L.~Pishchulin, S.~Wuhrer, T.~Helten, C.~Theobalt, B.~Schiele, Building
  statistical shape spaces for 3d human modeling, Pattern Recognition 67 (2017)
  276 -- 286.
\newblock \href {https://doi.org/https://doi.org/10.1016/j.patcog.2017.02.018}
  {\path{doi:https://doi.org/10.1016/j.patcog.2017.02.018}}.

\bibitem{Toshev_CVPR_2014}
A.~Toshev, C.~Szegedy, {DeepPose: Human Pose Estimation via Deep Neural
  Networks}, in: {Computer Vision and Pattern Recognition (CVPR)}, 2014, pp.
  1653--1660.

\bibitem{Pishchulin_ICCV_2013}
L.~Pishchulin, M.~Andriluka, P.~V. Gehler, B.~Schiele, Strong appearance and
  expressive spatial models for human pose estimation, in: International
  Conference on Computer Vision (ICCV), 2013, pp. 3487--3494.

\bibitem{Ladicky_CVPR_2013}
L.~Ladicky, P.~H.~S. Torr, A.~Zisserman, Human pose estimation using a joint
  pixel-wise and part-wise formulation, in: Computer Vision and Pattern
  Recognition (CVPR), 2013.

\bibitem{VNect_SIGGRAPH2017}
D.~Mehta, S.~Sridhar, O.~Sotnychenko, H.~Rhodin, M.~Shafiei, H.-P. Seidel,
  W.~Xu, D.~Casas, C.~Theobalt,
  \href{http://gvv.mpi-inf.mpg.de/projects/VNect/}{Vnect: Real-time 3d human
  pose estimation with a single rgb camera}, Vol.~36, 2017.
\newblock \href {https://doi.org/10.1145/3072959.3073596}
  {\path{doi:10.1145/3072959.3073596}}.
\newline\urlprefix\url{http://gvv.mpi-inf.mpg.de/projects/VNect/}

\bibitem{Chen_2017_CVPR}
C.-H. Chen, D.~Ramanan, 3d human pose estimation = 2d pose estimation +
  matching, in: The IEEE Conference on Computer Vision and Pattern Recognition
  (CVPR), 2017.

\bibitem{Newell_ECCV_2016}
A.~Newell, K.~Yang, J.~Deng, {Stacked Hourglass Networks for Human Pose
  Estimation}, {European Conference on Computer Vision (ECCV)} (2016) 483--499.

\bibitem{Sun_2018_ECCV}
X.~Sun, B.~Xiao, F.~Wei, S.~Liang, Y.~Wei, Integral human pose regression, in:
  The European Conference on Computer Vision (ECCV), 2018.

\bibitem{SARAFIANOS20161}
N.~Sarafianos, B.~Boteanu, B.~Ionescu, I.~A. Kakadiaris, 3d human pose
  estimation: A review of the literature and analysis of covariates, Computer
  Vision and Image Understanding 152 (2016) 1 -- 20.
\newblock \href {https://doi.org/https://doi.org/10.1016/j.cviu.2016.09.002}
  {\path{doi:https://doi.org/10.1016/j.cviu.2016.09.002}}.

\bibitem{ATREVI2017389}
D.~F. Atrevi, D.~Vivet, F.~Duculty, B.~Emile, A very simple framework for 3d
  human poses estimation using a single 2d image: Comparison of geometric
  moments descriptors, Pattern Recognition 71 (2017) 389 -- 401.
\newblock \href {https://doi.org/https://doi.org/10.1016/j.patcog.2017.06.024}
  {\path{doi:https://doi.org/10.1016/j.patcog.2017.06.024}}.

\bibitem{Ionesc_ICCV_2011}
C.~Ionescu, F.~Li, C.~Sminchisescu, {Latent structured models for human pose
  estimation}, in: {International Conference on Computer Vision (ICCV)}, 2011,
  pp. 2220--2227.

\bibitem{Moll_CVPR_2014}
G.~Pons-Moll, D.~J. Fleet, B.~Rosenhahn, Posebits for monocular human pose
  estimation, in: 2014 IEEE Conference on Computer Vision and Pattern
  Recognition, 2014, pp. 2345--2352.
\newblock \href {https://doi.org/10.1109/CVPR.2014.300}
  {\path{doi:10.1109/CVPR.2014.300}}.

\bibitem{TekinKSLF16}
B.~Tekin, I.~Katircioglu, M.~Salzmann, V.~Lepetit, P.~Fua, Structured
  prediction of 3d human pose with deep neural networks, in: Proceedings of the
  British Machine Vision Conference 2016, {BMVC} 2016, York, UK, September
  19-22, 2016, 2016.

\bibitem{Li_2015_ICCV}
S.~Li, W.~Zhang, A.~B. Chan, Maximum-margin structured learning with deep
  networks for 3d human pose estimation, in: The IEEE International Conference
  on Computer Vision (ICCV), 2015.

\bibitem{Ionescu_CVPR_2014}
C.~Ionescu, J.~Carreira, C.~Sminchisescu, Iterated second-order label sensitive
  pooling for 3d human pose estimation, in: 2014 IEEE Conference on Computer
  Vision and Pattern Recognition, 2014, pp. 1661--1668.

\bibitem{insafutdinov17cvpr}
E.~Insafutdinov, M.~Andriluka, L.~Pishchulin, S.~Tang, E.~Levinkov, B.~Andres,
  B.~Schiele, Arttrack: {A}rticulated multi-person tracking in the wild, in:
  30th IEEE Conference on Computer Vision and Pattern Recognition (CVPR 2017),
  IEEE, Honolulu, HI, USA, 2017, pp. 1293--1301.
\newblock \href {https://doi.org/10.1109/CVPR.2017.142}
  {\path{doi:10.1109/CVPR.2017.142}}.

\bibitem{Agarwal}
A.~Agarwal, B.~Triggs, Recovering 3d human pose from monocular images, IEEE
  Transactions on Pattern Analysis and Machine Intelligence 28~(1) (2006)
  44--58.

\bibitem{zhou2016deep}
X.~Zhou, X.~Sun, W.~Zhang, S.~Liang, Y.~Wei, Deep kinematic pose regression,
  Computer Vision ECCV 2016 Workshops.

\bibitem{Pavlakos_2017_CVPR}
G.~Pavlakos, X.~Zhou, K.~G. Derpanis, K.~Daniilidis, Coarse-to-fine volumetric
  prediction for single-image 3{D} human pose, in: Proceedings of the IEEE
  Conference on Computer Vision and Pattern Recognition, 2017.

\bibitem{Popa_CVPR_2017}
A.~Popa, M.~Zanfir, C.~Sminchisescu,
  \href{http://arxiv.org/abs/1701.08985}{Deep multitask architecture for
  integrated 2d and 3d human sensing}, arXiv preprint arXiv:1701.08985
  abs/1701.08985.
\newline\urlprefix\url{http://arxiv.org/abs/1701.08985}

\bibitem{Martinez_2017}
J.~Martinez, R.~Hossain, J.~Romero, J.~J. Little, A simple yet effective
  baseline for 3d human pose estimation, in: ICCV, 2017.

\bibitem{Sun_2017_ICCV}
X.~Sun, J.~Shang, S.~Liang, Y.~Wei, Compositional human pose regression, arXiv
  preprint arXiv:1702.07432.

\bibitem{luo2018orinet}
C.~Luo, X.~Chu, A.~Yuille, Orinet: A fully convolutional network for 3d human
  pose estimation, in: BMVC, 2018.

\bibitem{Mehta_2017_3DV}
D.~Mehta, H.~Rhodin, D.~Casas, P.~Fua, O.~Sotnychenko, W.~Xu, C.~Theobalt,
  \href{http://gvv.mpi-inf.mpg.de/3dhp_dataset}{Monocular 3d human pose
  estimation in the wild using improved cnn supervision}, in: 3D Vision (3DV),
  2017 Fifth International Conference on, 2017.
\newline\urlprefix\url{http://gvv.mpi-inf.mpg.de/3dhp_dataset}

\bibitem{Wei_CVPR_2016}
S.-E. Wei, V.~Ramakrishna, T.~Kanade, Y.~Sheikh, {Convolutional pose machines},
  in: {IEEE Conference on Computer Vision and Pattern Recognition (CVPR)},
  2016.

\bibitem{Gkioxari_ECCV_2016}
G.~Gkioxari, A.~Toshev, N.~Jaitly, {Chained Predictions Using Convolutional
  Neural Networks}, European Conference on Computer Vision (ECCV).

\bibitem{Hu_2016_CVPR}
P.~Hu, D.~Ramanan, Bottom-up and top-down reasoning with hierarchical rectified
  gaussians, in: The IEEE Conference on Computer Vision and Pattern Recognition
  (CVPR), 2016.

\bibitem{Bulat_ECCV_2016}
A.~Bulat, G.~Tzimiropoulos, {Human pose estimation via Convolutional Part
  Heatmap Regression}, in: European Conference on Computer Vision (ECCV), 2016,
  pp. 717--732.

\bibitem{Zhou_2017_ICCV}
X.~Zhou, Q.~Huang, X.~Sun, X.~Xue, Y.~Wei, Towards 3d human pose estimation in
  the wild: A weakly-supervised approach, in: The IEEE International Conference
  on Computer Vision (ICCV), 2017.

\bibitem{Luvizon_2017_CoRR}
D.~C. Luvizon, H.~Tabia, D.~Picard,
  \href{http://arxiv.org/abs/1710.02322}{Human pose regression by combining
  indirect part detection and contextual information}, CoRR abs/1710.02322.
\newblock \href {http://arxiv.org/abs/1710.02322} {\path{arXiv:1710.02322}}.
\newline\urlprefix\url{http://arxiv.org/abs/1710.02322}

\bibitem{Chollet_2017_CVPR}
F.~Chollet, Xception: Deep learning with depthwise separable convolutions, in:
  The IEEE Conference on Computer Vision and Pattern Recognition (CVPR), 2017.

\bibitem{HowardZCKWWAA17}
A.~G. Howard, M.~Zhu, B.~Chen, D.~Kalenichenko, W.~Wang, T.~Weyand,
  M.~Andreetto, H.~Adam, Mobilenets: Efficient convolutional neural networks
  for mobile vision applications, CoRR abs/1704.04861.

\bibitem{h36m_pami}
C.~Ionescu, D.~Papava, V.~Olaru, C.~Sminchisescu, Human3.6m: Large scale
  datasets and predictive methods for 3d human sensing in natural environments,
  TPAMI 36~(7) (2014) 1325--1339.

\bibitem{Andriluka_CVPR_2014}
M.~Andriluka, L.~Pishchulin, P.~Gehler, B.~Schiele, {2D Human Pose Estimation:
  New Benchmark and State of the Art Analysis}, in: {IEEE Conference on
  Computer Vision and Pattern Recognition (CVPR)}, 2014.

\bibitem{Zou05regularizationand}
H.~Zou, T.~Hastie, Regularization and variable selection via the elastic net,
  Journal of the Royal Statistical Society, Series B 67 (2005) 301--320.

\end{thebibliography}

\end{document}